\documentclass[a4, 10 pt, conference]{ieeeconf}  % Comment this line out if you need a4paper

\IEEEoverridecommandlockouts                              % This command is only needed if 
\usepackage{graphicx} % for pdf, bitmapped graphics files
\usepackage{balance}

\title{\LARGE \bf
Overview of Dialogue Robot Competition 2022
}

\author{Takashi Minato$^{1,2}$, Ryuichiro Higashinaka$^{3}$, Kurima Sakai$^{1}$, Tomo Funayama$^{1}$,\\ Hiromitsu Nishizaki$^{4}$, and Takayuki Nagai$^{5}$ %
\thanks{$^{1}$Hiroshi Ishiguro Laboratories, ATR, Kyoto, Japan
        {\tt\small \{minato, kurima.sakai, funayama\}@atr.jp}}%
\thanks{$^{2}$Guardian Robot Project, RIKEN, Kyoto, Japan
        {\tt\small takashi.minato@riken.jp}}%
\thanks{$^{3}$Graduate School of Informatics, Nagoya University, Nagoya, Japan
        {\tt\small higashinaka@i.nagoya-u.ac.jp}}%
\thanks{$^{4}$Graduate Faculty of Interdisciplinary Research, University of Yamanashi, Yamanashi, Japan
        {\tt\small hnishi@yamanashi.ac.jp}}%        
 \thanks{$^{5}$Graduate School of Engineering Science, Osaka University, Osaka, Japan
        {\tt\small nagai@sys.es.osaka-u.ac.jp}}%          
}

\begin{document}

\maketitle
\thispagestyle{empty}
\pagestyle{empty}

%%%%%%%%%%%%%%%%%%%%%%%%%%%%%%%%%%%%%%%%%%%%%%%%%%%%%%%%%%%%%%%%%%%%%%%%%%%%%%%%
\begin{abstract}
Although many competitions have been held on dialogue systems in the past, no competition has been organized specifically for dialogue with humanoid robots. As the first such attempt in the world, we held a dialogue robot competition in 2020 to compare the performances of interactive robots using an android that closely resembles a human. Dialogue Robot Competition 2022 (DRC2022) was the second competition, held in August 2022. The task and regulations followed those of the first competition, while the evaluation method was improved and the event was internationalized. The competition has two rounds, a preliminary round and the final round. In the preliminary round, twelve participating teams competed in performance of a dialogue robot in the manner of a field experiment, and then three of those teams were selected as finalists. The final round will be held on October 25, 2022, in the Robot Competition session of IROS2022. This paper provides an overview of the task settings and evaluation method of DRC2022 and the results of the preliminary round. 
\end{abstract}

%%%%%%%%%%%%%%%%%%%%%%%%%%%%%%%%%%%%%%%%%%%%%%%%%%%%%%%%%%%%%%%%%%%%%%%%%%%%%%%%
\section{INTRODUCTION}

Voice interactive devices used in our daily lives have been developed from the automated answering services for telephones in the past to the voice agents and AI speakers of smartphones in recent years, and humanoid interactive robots have become the focus of developing next-generation voice interactive devices \cite{Inoue2016}. To promote their development, we have organized a dialogue robot competition. We expect this competition to lead to the development of robust spoken dialogue technology that can improve user satisfaction.

Although there have been many competitions for dialogue systems in the past \cite{Khatri2018,Higashinaka2021}, no competition specifically for dialogue with humanoid robots has been organized. If humanoid robots are used for dialogue services, it is expected that more user-friendly and hospitable services \cite{Collins2020} could be achieved by using various modalities such as robot's facial expressions and gestures. Therefore, as the world's first such attempt, we held a dialogue robot competition in 2020 to compare the performances of interactive robots using an android that closely resembles a human. The competition is remarkable in that it uses an android robot commonly shared by all participating teams, which are provided with middleware to control the robot as well as minimal software for speech recognition and facial recognition of the person interacting with the robot. This allows the participating teams to focus on the development of the dialogue management module, which is the core of a dialogue system.

Eleven teams participated in the first competition, and the best team was selected in the final round in October 2021. A variety of dialogue strategies were proposed, and the competition demonstrated that hospitality systems could be created using dialogue robots \cite{GCCE2022}. 

Following this success, we organized a second competition, Dialogue Robot Competition 2022 (hereafter, DRC2022). In DRC2022, we followed the dialogue task of the first competition but improved the procedure for evaluating the dialogue systems. We performed a two-stage evaluation as done in the previous competition: first, a preliminary round in which the systems of the participating teams were evaluated by the general public, and next, a final round in which the systems were evaluated by designated dialogue researchers to evaluate the technical aspects and by experts working in the tourism industry to evaluate the customer service performance. In DRC2022, we chose the National Museum of Emerging Science and Innovation (Miraikan), which is visited by many people interested in science and technology, as the venue for the preliminary round. In addition, we decided to hold the final round in the Robot Competition session of IROS2022, which is the largest international robotics conference, to share our activities with robotics researchers throughout the world and to serve as a stepping stone toward internationalization of the competition. DRC2022 was supported by JTB Corporation and JTB Publishing Inc. as in the previous competition. This paper gives an outline of DRC2022 and the results of the preliminary round.

\section{TASK SETTINGS}
The tasks and regulations are the same as those for the first competition. In this task, named the travel destination recommendation task, the robot works as a counter salesperson and assists the customer in deciding on a sightseeing spot in the region where the travel agency is based. Before the dialogue, the customer selects two places to visit as candidates, from six possible spots, and then decides where to visit among the candidates through a dialogue with the robot. In order to properly assist the customer, the robot needs to explain the highlights of the two spots, answer their queries, and help them decide which of the two spots to visit. The participating teams aim to develop a spoken dialogue robot system that enables customers to have a satisfying conversation with the robot and eventually choose the destination that the robot tries to recommend. The participating teams were provided with information about the sightseeing spots (JTB's "Rurubu DATA"), but they were also allowed to use external resources such as the Web for more informative utterances. The duration of each dialogue was set to approx. 5 minutes, and it was conducted in Japanese .

\section{AVAILABLE RESOURCES}
\begin{figure}[tb]
    \centering
    \includegraphics[scale=0.105]{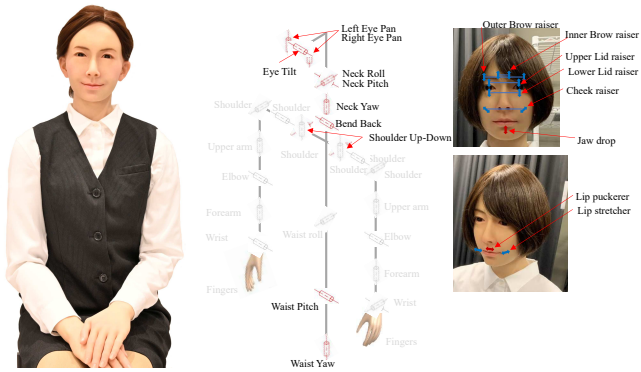}
    \caption{Android I used in the competition}
    \label{fig:android}
\end{figure}

\begin{figure}[tb]
    \centering
    \includegraphics[scale=0.1]{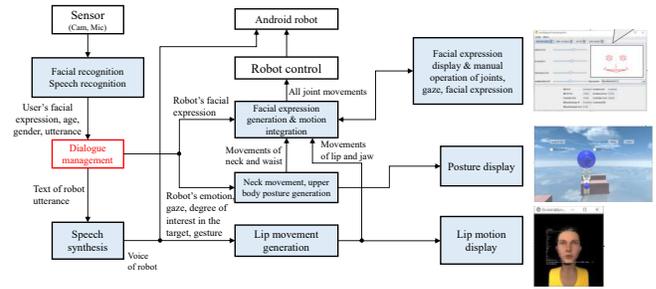}
    \caption{Overview of the middleware distributed to participating teams}
    \label{fig:middleware}
\end{figure}

The Android I was made available as hardware; it is approximately 165 cm tall and has a feminine
appearance. Its face and hands are covered with soft silicone skin. It has 18 degrees of freedom (Fig.~\ref{fig:android}) and is capable of lip movements synchronized with speech, head movements such as nodding, eye blinking, gaze behavior, facial expressions, and postural changes of the upper body, all driven by pneumatic actuators. Neither of the arms is movable. In addition, participating teams were provided with middleware that integrates face recognition, speech recognition and synthesis, robot facial expression generation, gaze control, lip movement generation, head movement generation, and posture control (Fig.~\ref{fig:middleware}). The dialogue management module developed by the participating teams simply inputs the customer's facial recognition results (facial expression, gender and age), along with the content of the customer's speech, and outputs utterances with the robot's multimodal expressions such as facial expressions, gestures, and body movements as additional information. The middleware comes with a robot simulator, allowing development without the need for an actual robot. A remote testing environment has also been developed so that the participating teams can remotely verify the actual behavior of the robot.

\section{PRELIMINARY ROUND}
\begin{figure}[tb]
    \centering
    \includegraphics[scale=0.15]{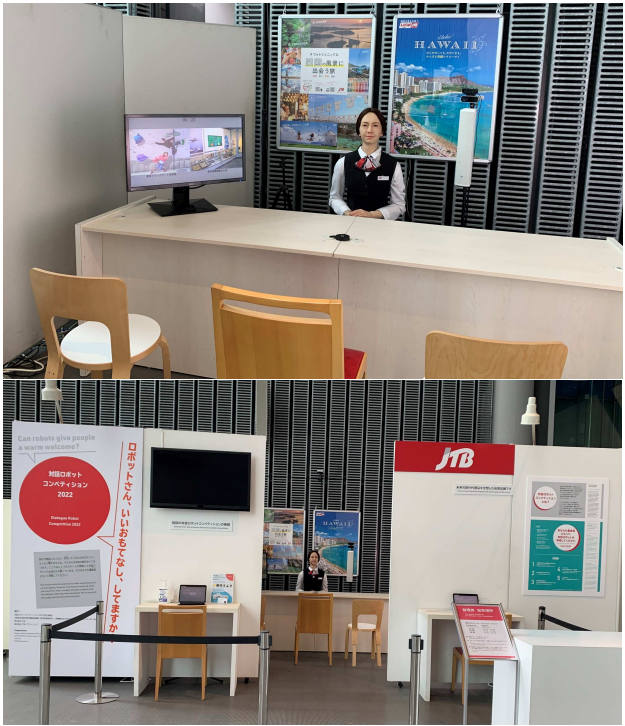}
    \caption{Mock travel agency at Miraikan}
    \label{fig:travel_agency}
\end{figure}

The preliminary round was held August 11--27, 2022, as a part of the exhibition "From Imagination to Implementation: Picturing Our Future with Robots \verb|<|Miraikan\verb|>|." A mock travel agency was set up in the Symbol Zone of Miraikan (Fig.~\ref{fig:travel_agency}), and the visitors interested in talking with the robot had conversation with it while acting as customers (considering the content of the dialogue, customers were limited to 12 years of age or older). Tables, chairs, counter salesperson uniforms, and logos actually used in travel agencies provided a more realistic simulated travel agency. The visitors who wished to talk with the robot received an explanation of the dialogue situation from a staff member, selected two desired sightseeing spots, and then interacted as a customer with the robot. The six possible spots were:
\begin{itemize}
    \item Madame Tussauds Tokyo
    \item Tokyo Trick Art Museum
    \item Tokyo Water Science Museum
    \item Tokyo Customs Information Hiroba
    \item Telecom Center Observatory Lounge View Tokyo
    \item The Tokyo Rinkai Disaster Prevention Park
\end{itemize}
The customer selected two of the six spots before the dialogue. The spot that the robot wanted to recommend was decided by randomly choosing one of the two spots. After the dialogue, the visitors were asked to evaluate their impression of the dialogue. Due to the spread of COVID-19, it was assumed that the visitor would talk with the robot while wearing a mask. In such a case, the gender and age estimation by the facial recognition system could not work well. To compensate for this issue, we asked the visitor to answer his/her age and gender before the dialogue as well as the two desired spots, so that the robot could start the dialogue with this information. Considering the accuracy of the age estimation by the facial recognition system, the visitor's age is handled as categorical information like early teens, late teens, early 20s, late 20s, etc.

The dialogue performance was evaluated by two factors: impression evaluation
and robot recommendation effect. The impression evaluation consisted of the
following nine questions (originally in Japanese; translated here into English).
\begin{itemize}
    \item Satisfaction with choice: "Were you satisfied with your choice of tourist attraction to visit?"
    \item Informativeness: "Were you able to obtain sufficient information about the sightseeing spots?"
    \item Naturalness: "Did you have a natural dialogue with the robot?"
    \item Appropriateness: "Was the robot's service appropriate?"
    \item Likeability: "Was the robot likable in providing the service?"
    \item Satisfaction with dialogue: "Were you satisfied with your interaction with the robot?"
    \item Trustworthiness: "Did you trust the robot?"
    \item Usefulness: "Did you use the information obtained from the robot to select the sightseeing spot?"
    \item Intention to reuse: "Would you like to visit this travel agency again?"
\end{itemize}

\begin{figure}[tb]
    \centering
    \includegraphics[scale=0.105]{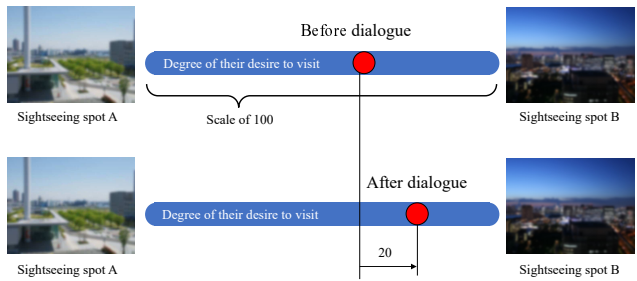}
    \caption{Interface for evaluating the effect of robot recommendation. If the robot's recommended spot is sightseeing spot B, the robot recommendation effect is 20 in the case of the above illustration. The effect is -20 if the recommended spot is sightseeing spot A.}
    \label{fig:interface}
\end{figure}

Each item was measured on a 7-point Likert scale. The effectiveness of the robot's recommendation was evaluated by the change in the degree to which the customer wanted to visit the sightseeing spot recommended by the robot before and after the dialogue. The customers were not informed of which destination was recommended by the robot. They responded to the degree of their desire to visit either of the two sightseeing spots (on a scale of 100) before and after the dialogue, using the interface shown in Fig.~\ref{fig:interface}. The increase in the degree of desire to go to the recommended sightseeing spot was considered to be the recommendation effect (this value is negative if the degree decreases).

To comprehensively evaluate the two factors of impression and recommendation effect, the two evaluation scores (total of averaged impression scores and averaged robot's recommendation effect score) for each participating team were plotted on a scatter plot with two axes. In this plot, a team belonging to the cluster formed at the position with the highest values on both axes was considered a top team of preliminary round and advanced to the final round. The teams belonging to the cluster formed at the second highest position received an honorable mention award.

\begin{figure}[tb]
    \centering
    \includegraphics[scale=0.095]{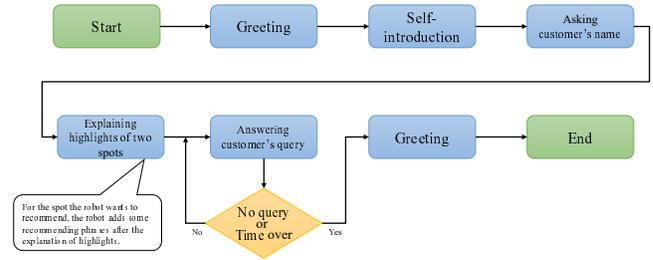}
    \caption{Dialogue flow of baseline system}
    \label{fig:baseline}
\end{figure}

To provide a baseline for evaluating the dialogue systems of the participating teams, we developed a baseline dialogue system. The baseline system follows the scenario shown in Fig.~\ref{fig:baseline}. The robot basically explains the highlights of the two sightseeing spots selected by the customer in advance and then answers the customers' queries about the spots if they have them. In order for the robot to limit its talk to the information about the sightseeing spots given in advance, the following restrictions are imposed on the baseline system:
\begin{itemize}
\item No information about the sightseeing spots is obtained from the Web.
\item The robot answers the customer's query by using only the information given in advance. The robot simply replies, "I don't know" or "I can't answer" if an answer to the query cannot be found in the given information.
\item The robot does not use the customer's age or gender information and does not try to discern the customer's preference through the dialogue.
\item The robot does not produce any ice breaker behavior or small talk.
\end{itemize}
In addition, the robot's nonverbal behaviors such as gaze and interval taking are developed based on heuristic rules to ensure natural behavior.

\begin{figure}[tb]
    \centering
    \includegraphics[scale=0.25]{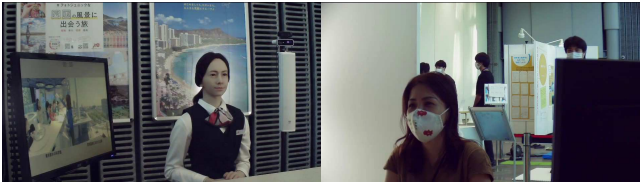}
    \caption{Example scene of dialogue in the preliminary round}
    \label{fig:scene}
\end{figure}

\begin{figure}[tb]
    \centering
    \includegraphics[scale=0.20]{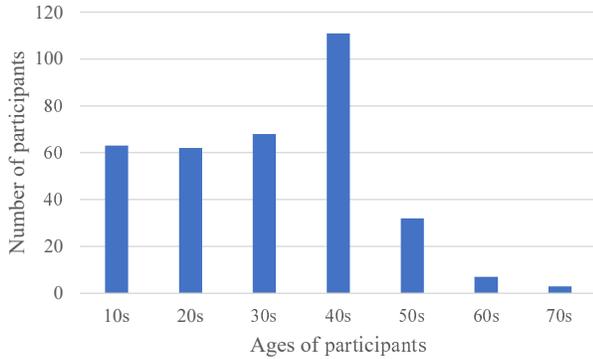}
    \caption{Age distribution of customers}
    \label{fig:age}
\end{figure}

Fourteen teams submitted entries to DRC2022, and finally twelve teams (CIS, ditlab, DSML-TDU, Flow, irisapu, ISC22, LINE, MIYABI, MIYAMA, OS, ponponkichi, SZK-L) participated in the preliminary round along with the baseline system. Eight of the teams were from universities, one from a college of technology, one from a vocational college, and two from industry. A total of 347 customers (174 males, 169 females, and 4 not identified) evaluated the system in the preliminary round. Fig.~\ref{fig:scene} shows a scene of a customer talking with the robot. The average number of customers handled per team was 26.7. From the age distribution of customers shown in Fig.~\ref{fig:age}, it can be seen that a wide range of generations, from teens to people in their 70s, talked with the robot. 

\begin{figure}[tb]
    \centering
    \includegraphics[scale=0.1]{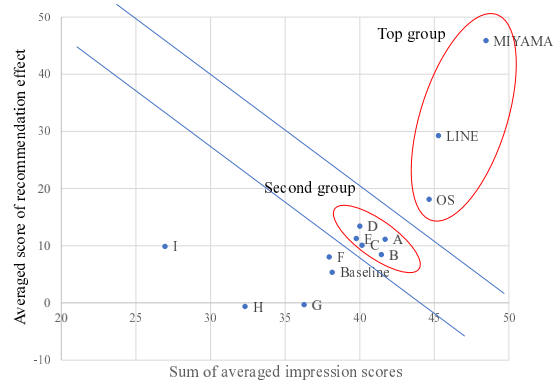}
    \caption{Scatter plot with impression and recommendation scores}
    \label{fig:plot}
\end{figure}

The evaluation results are shown in Table~\ref{table:result}, and the scatter plot is shown in Fig.~\ref{fig:plot}. Teams that did not make it to the final round are anonymized as "Team X." Since the scores of three teams (MIYAMA, LINE, and OS) significantly outperformed those of the other teams, these three teams were selected as finalists.

\begin{table*}[tb]
    \caption{Results of preliminary round. The teams are sorted by total impression score. Sat/c, Inf, Nat, App, Lik, Sat/d, Tru, Use, Reu, and Recom denote Satisfaction with choice, Informativeness, Naturalness, Appropriateness, Likeability, Satisfaction with dialogue, Trustworthiness, Usefulness, Intention to reuse, and Robot recommendation effect, respectively.}
    \label{table:result}
    \begin{tabular}{c} 
    \begin{minipage}{180mm}
      \centering
      \scalebox{0.18}{\includegraphics{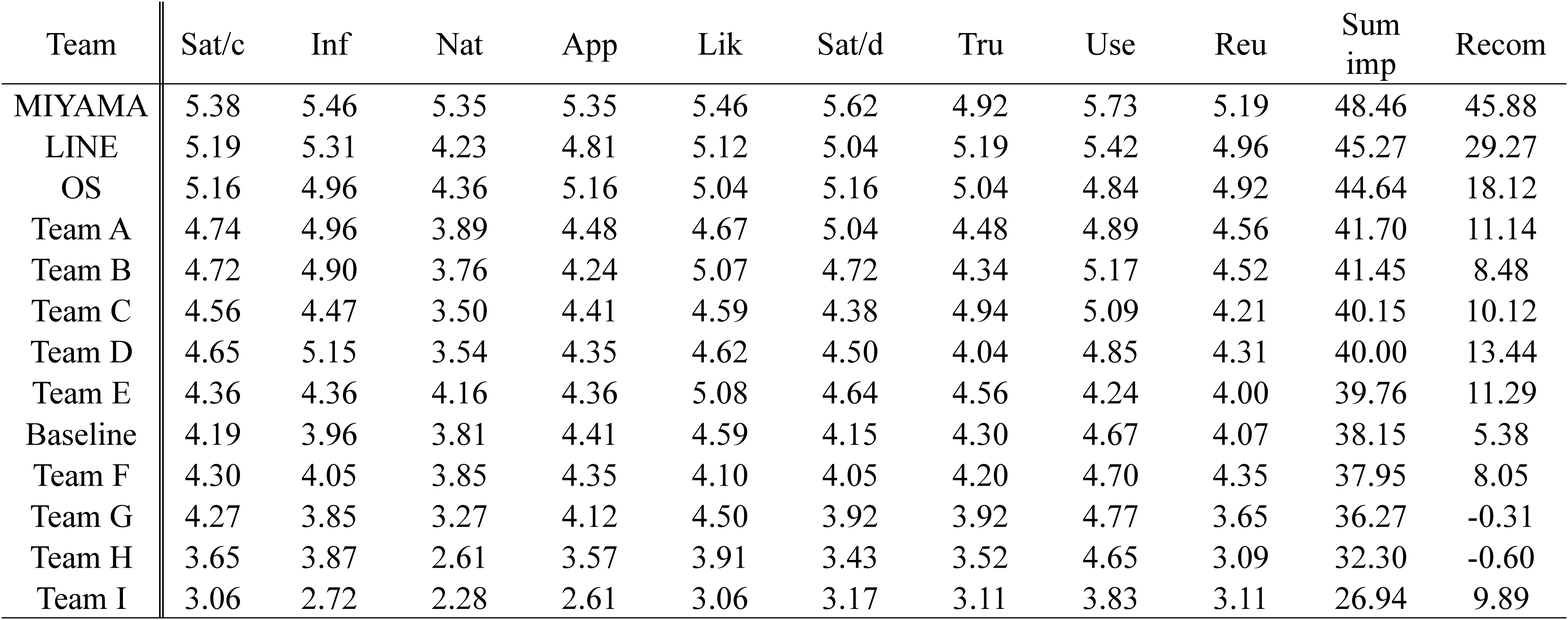}}
    \end{minipage} 
    \end{tabular}
\end{table*}

\begin{figure}[tb]
    \centering
    \includegraphics[scale=0.16]{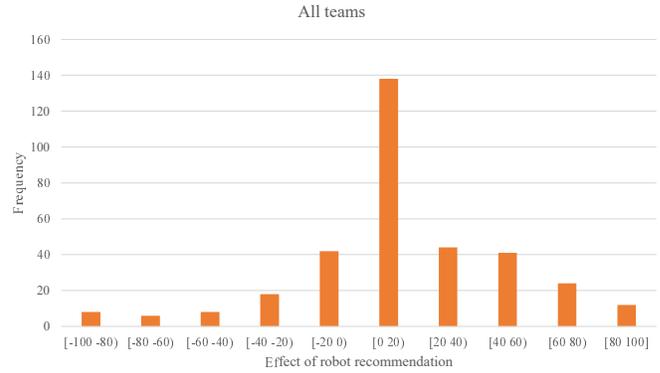}
    \caption{Histogram of scores of robot recommendation effect (all teams)}
    \label{fig:histogram_all}
\end{figure}

\begin{figure}[tb]
    \centering
    \includegraphics[scale=0.16]{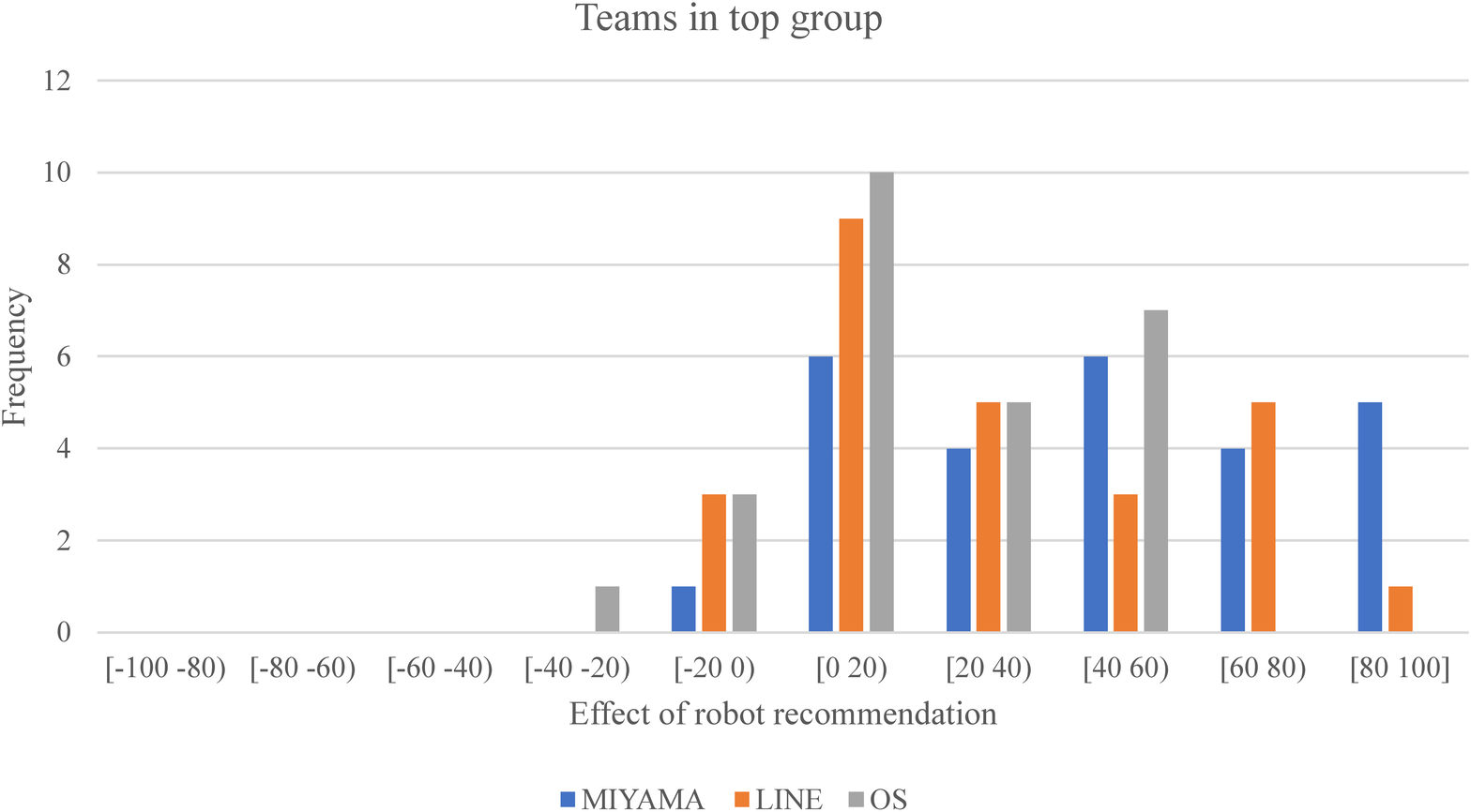}
    \caption{Histogram of scores of robot recommendation effect (three teams in top group)}
    \label{fig:histogram_top}
\end{figure}

The top-three teams showed a significant difference in their scores of robot recommendation effect from those of the other teams. For most teams, the recommendation effect scores are widely distributed between -100 and 100. Fig.~\ref{fig:histogram_all} shows the distribution of recommendation effect scores from all 12 teams and the Baseline. Some customers increased their desire to visit the sightseeing spot recommended by the robot, while others wanted to visit the spot that was not recommended. On the other hand, the distribution of scores for the top three teams (Fig.~\ref{fig:histogram_top}) show that most of the customers of those teams increased their desire to visit the robot's recommended sightseeing spot. In other words, the robot's recommendation had a strong influence on the customers' choice of sightseeing spots. From this viewpoint, we can conclude  that the top three teams achieved better performance than the other teams.

Teams A, B, C, D, and E form the second group in Fig.~\ref{fig:plot}. Those five teams received an honorable mention award. In the previous competition, Baseline was 4th place among the twelve teams, but more than half of the teams outperformed the baseline score in DRC2022. Therefore, the overall performance of the dialogue robot systems developed by the participating teams has improved.

\section{OVERVIEW OF DIALOGUE SYSTEMS DEVELOPED BY PARTICIPATING TEAMS}

Most of the teams designed a dialogue flow to accomplish the travel destination recommendation task. Basically, the dialogue is designed to follow a scenario where the robot makes a self-introduction, asks the customer's preference, introduces two candidates of sightseeing spots, answers the customer's questions about the sightseeing spots, and recommends one of two sightseeing spots.

The task requires user-friendly and hospitable services; therefore, many teams implemented dialogue strategies to adapt the robot to the customer's traits. For example:
\begin{itemize}
\item The robot estimates the customer's personality based on the face image at the beginning of the dialogue and adaptively changes the inquiry patterns based on the estimated personality (team MIYAMA).
\item The robot recognizes a sentiment of the customer's utterance and tries to empathize with his or her positive utterance (team OS).
\item The robot recognizes a fashion item worn/carried by the customer and makes an icebreaker utterance that mentions it (team DSML-TDU).
\item The robot adaptively changes its speech style based on the customer's age (teams irisapu and ditlab).
\end{itemize}

The robot's nonverbal expressions such as gaze and facial expression rather
than verbal expressions are effective in creating a comfortable interaction,
which is different from spoken dialogue systems. A number of teams
implemented the robot's nonverbal expressions for that purpose. For example:
\begin{itemize}
\item The robot expresses an emotion by controlling the speech rate, voice pitch, facial expression, and body posture (team OS).
\item The robot makes facial expressions and gaze behaviors as a way to create a comfortable interaction (team MIYABI).
\end{itemize}
As a unique attempt, team MIYABI proposed a dialogue strategy to elicit the customer's pointing gesture. It is expected that customers would feel a strong sense of joint participation with the robot by using bodily gestures in the interaction.

The dialogue is task-oriented, but open-domain questions are assumed when a customer wants to know about the sightseeing spots. Many teams designed robot-initiative dialogue flows and used closed-style questions to avoid an open-domain dialogue. A few teams tried to integrate a rule-based dialogue system and an open-domain dialogue system. Team LINE integrated a rule-based system following a pre-designed scenario and an originally developed open-domain dialogue system called HyperCLOVA. The impression scores in the preliminary round were high, showing that the system worked successfully. The robot could correctly answer various questions from the customers, and this might improve the impression of the robot's reliability as a specialist of the travel agency. Team ditlab also integrated rule-based and open-domain dialogue systems. In addition, the system of ditlab attempted to correctly recognize a customer's name using a Japanese-name database. In service businesses, correctly recognizing the customer's name is important. Accordingly, the LINE and ditlab teams are attempting to develop a practical dialogue system.

Team SZK-L proposed a method to decide the recommendation of sightseeing spot in a quantitative manner. It vectorizes a customer's preference for sightseeing and the attributes of sightseeing spots and then quantitatively assesses the closeness between customer preference and sightseeing spots attributes. Team ditlab's system also quantitatively assesses which sightseeing spot matches a customer's preference. It refers to attribute information on the web and covers the customer's various preferences.

Teams MIAMA and LINE introduced their original recognition systems in addition to the system provided by the competition organizer. Team MIYAMA developed a system to estimate the customer's personality based on the facial image and to adapt the dialogue strategy to the customer's personality. Team LINE developed a speech recognition system.

As an attempt to improve the architecture of the robot's dialogue system, team Flow designed a system with a pipeline composed of natural language understanding (NLU), dialogue state tracking (DST), and natural language generation (NLG) modules with the aim of optimizing dialogue performance. This system has the potential to integrate all modules and optimize the performance of the entire system. In the preliminary round, the optimization did not work, but this is still a promising approach to making a robust dialogue system in the future.

The top-three systems (MIYAMA, LINE, and OS) in the preliminary round are distinctive in their adaptability to customer's personality, reliability as a specialist of the travel agency, and empathetic interaction. The robot dialogue systems with not only accurate information conveyance to the customer but also value-added service operations were highly rated in the preliminary round.

\section{FINAL ROUND}
The final round will be held on October 25, 2022, in the Robot Competition session of IROS2022. In the final round, the systems will be evaluated by designated dialogue researchers and by experts working in the tourist industry. The dialogue will be open to researchers watching the Robot Competition (dialogue will be in Japanese, but text translated into English will be displayed in real time).

\section{CONCLUSION}
This paper provided an overview of DRC2022 and the results of the preliminary round. The top-three teams selected as finalists significantly outperformed the other nine teams in both of the key factors: impression evaluation and robot recommendation effect. The baseline system was almost the same as that used in the previous competition, and thus its lowered ranking among the competitors suggests that the overall performance of the dialogue robot systems developed by the participating teams in this competition improved over that in the previous competition.

\balance
%\addtolength{\textheight}{-4cm}   % This command serves to balance the column lengths

%%%%%%%%%%%%%%%%%%%%%%%%%%%%%%%%%%%%%%%%%%%%%%%%%%%%%%%%%%%%%%%%%%%%%%%%%%%%%%%%

%%%%%%%%%%%%%%%%%%%%%%%%%%%%%%%%%%%%%%%%%%%%%%%%%%%%%%%%%%%%%%%%%%%%%%%%%%%%%%%%

%%%%%%%%%%%%%%%%%%%%%%%%%%%%%%%%%%%%%%%%%%%%%%%%%%%%%%%%%%%%%%%%%%%%%%%%%%%%%%%%

\section*{ACKNOWLEDGMENT}

This work was supported by a Grant-in-Aid for Scientific Research (Grant No. JP19H05692).

%%%%%%%%%%%%%%%%%%%%%%%%%%%%%%%%%%%%%%%%%%%%%%%%%%%%%%%%%%%%%%%%%%%%%%%%%%%%%%%%

\bibliographystyle{IEEEtran}
\bibliography{drc2022}

\end{document}